\documentclass[conference]{IEEEtran}
\usepackage{cite}
\usepackage{amsmath,amssymb,amsfonts}
\usepackage{graphicx}
\usepackage{booktabs}
\usepackage{multirow}
\usepackage{stfloats}
\usepackage{graphicx}
\usepackage{tikz}
\providecommand{\ATPOsScore}[2]{%
  \makebox[12mm][c]{%
    \if\relax\detokenize{#1}\relax
      \ensuremath{\text{---}\,{\scriptscriptstyle\pm}\,\text{---}}%
    \else
      \ensuremath{#1\,{\scriptscriptstyle\pm #2}}%
    \fi}}

\def\BibTeX{{\rm B\kern-.05em{\sc i\kern-.025em b}\kern-.08em
    T\kern-.1667em\lower.7ex\hbox{E}\kern-.125emX}}

\usepackage{xcolor}
\definecolor{DeepText}{HTML}{202428}
\definecolor{DeepMuted}{HTML}{65737B}
\definecolor{DeepAccent}{HTML}{075EAB}
\definecolor{DeepPanelGray}{HTML}{F0F1F2}
\definecolor{DeepRule}{HTML}{292D32}
\usepackage[colorlinks=true,urlcolor=DeepAccent]{hyperref}
\makeatletter
\newcommand{\makebrandheader}{%
  \noindent
  \begin{minipage}[b]{0.68\textwidth}%
    \includegraphics[width=1.80in,trim=99bp 258bp 83bp 260bp,clip]{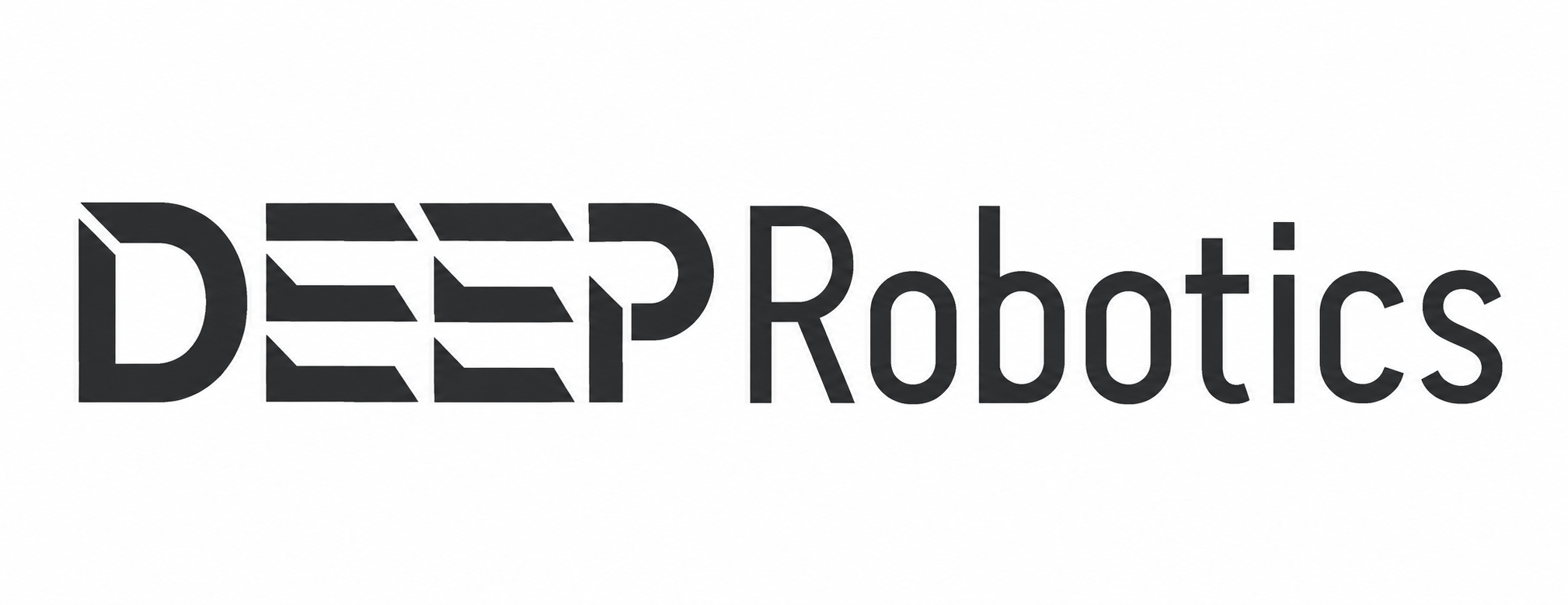}%
  \end{minipage}\hfill
  \begin{minipage}[b]{0.20\textwidth}%
    \raggedleft
    \includegraphics[height=0.50in,trim=50bp 3bp 58bp 0bp,clip]{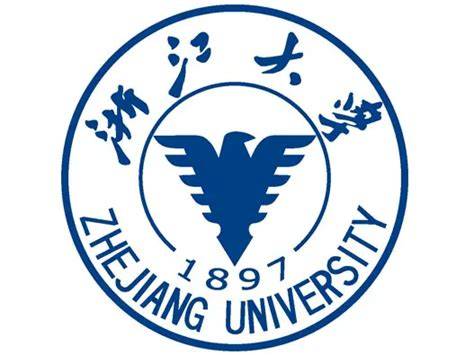}%
  \end{minipage}\par
  \vspace{0.08in}%
  {\color{DeepRule}\hrule height 0.8pt}%
  \vspace{0.12in}%
}
\newsavebox{\deep@frontmatterbox}
\newenvironment{deeprobotfrontmatter}{%
  \par\noindent
  \begin{lrbox}{\deep@frontmatterbox}%
  \begin{minipage}{\dimexpr\textwidth-0.48in\relax}%
  \setlength{\parindent}{0pt}%
  \setlength{\parskip}{0.48em plus 0.08em minus 0.05em}%
  \renewenvironment{abstract}{%
    \par\vspace{1.25em}%
    \normalfont\small\color{DeepText}\noindent\ignorespaces
  }{\par}%
}{%
  \end{minipage}%
  \end{lrbox}%
  \begingroup
    \setlength{\fboxsep}{0.24in}%
    \noindent\colorbox{DeepPanelGray}{\usebox{\deep@frontmatterbox}}%
  \endgroup
  \par\vspace{0.22in}%
}
\newcommand{\makefrontmattertitle}{%
  {\fontsize{21}{24}\selectfont\rmfamily\bfseries\color{DeepText}\@title\par}%
  \vspace{0.55em}%
  {\large\sffamily\mdseries\color{DeepText}\@author\par}%
  \vspace{0.22em}%
  {\normalsize\sffamily\color{DeepMuted}%
    \textsuperscript{1}DEEP Robotics, Hangzhou, China\quad
    \textsuperscript{2}Zhejiang University, Hangzhou, China\par}%
  \vspace{0.22em}%
  {\small\sffamily\color{DeepMuted}%
    \textsuperscript{\dag} Corresponding authors.\quad
    \textsuperscript{*} Project leader.\par}%
}
\makeatother

\begin{document}

\title{DR-APO: Decoupled-Representation\\
Advantage-Guided Policy Optimization}
\author{Runhua Zhang\textsuperscript{1,2}\quad
Yingjie Niu\textsuperscript{1,\dag,*}\quad
Chao Li\textsuperscript{1,\dag}}

\twocolumn[{%
\makebrandheader
\begin{deeprobotfrontmatter}
\makefrontmattertitle

\textit{Project page:}
\href{https://developer.deeprobotics.cn/research/articles/dr-apo/index.html}
{developer.deeprobotics.cn/research/articles/dr-apo}

\begin{abstract}
Policy representations in reinforcement learning 
have expanded beyond Gaussian policies to 
diffusion and flow-matching based generative policies.
However, existing RL methods remain representation-specific, lacking a unified
optimization interface for emerging generative policies.
In particular, extending likelihood-based objectives to multi-step generative
policies couples optimization to their internal generation process, where
accumulated likelihood ratios can weaken the update signal.
Meanwhile, generative-policy-specific methods often rely
on multi-candidate generation and critic-based selection,
improving overall return while providing limited improvement
to the policy itself.
We introduce \textbf{DR-APO
(Decoupled-Representation Advantage-Guided Policy Optimization)},
which decouples policy improvement from policy representation through
\emph{advantage-guided target improvement} and
\emph{representation-compatible direct action fitting},
providing a shared optimization principle for Gaussian, diffusion, and flow policies.
Across five MuJoCo-v5 tasks,
DR-APO remains competitive with SAC for Gaussian policies
and achieves the best aggregate policy-only performance
for both diffusion and flow-matching policies.
NavDP post-training improves the Success Rate (SR) from 70.33\% to 89.55\%
in simulation and from 25\% to 65\% in real-world navigation.
These results demonstrate a unified policy-improvement principle across
Gaussian, diffusion, and flow policies, from continuous-control learning to
embodied-policy post-training.
\end{abstract}
\end{deeprobotfrontmatter}
}]



\begin{figure}[!t]
\centering
\includegraphics[
    width=\columnwidth,
    trim=0 9mm 0 0,
    clip
]{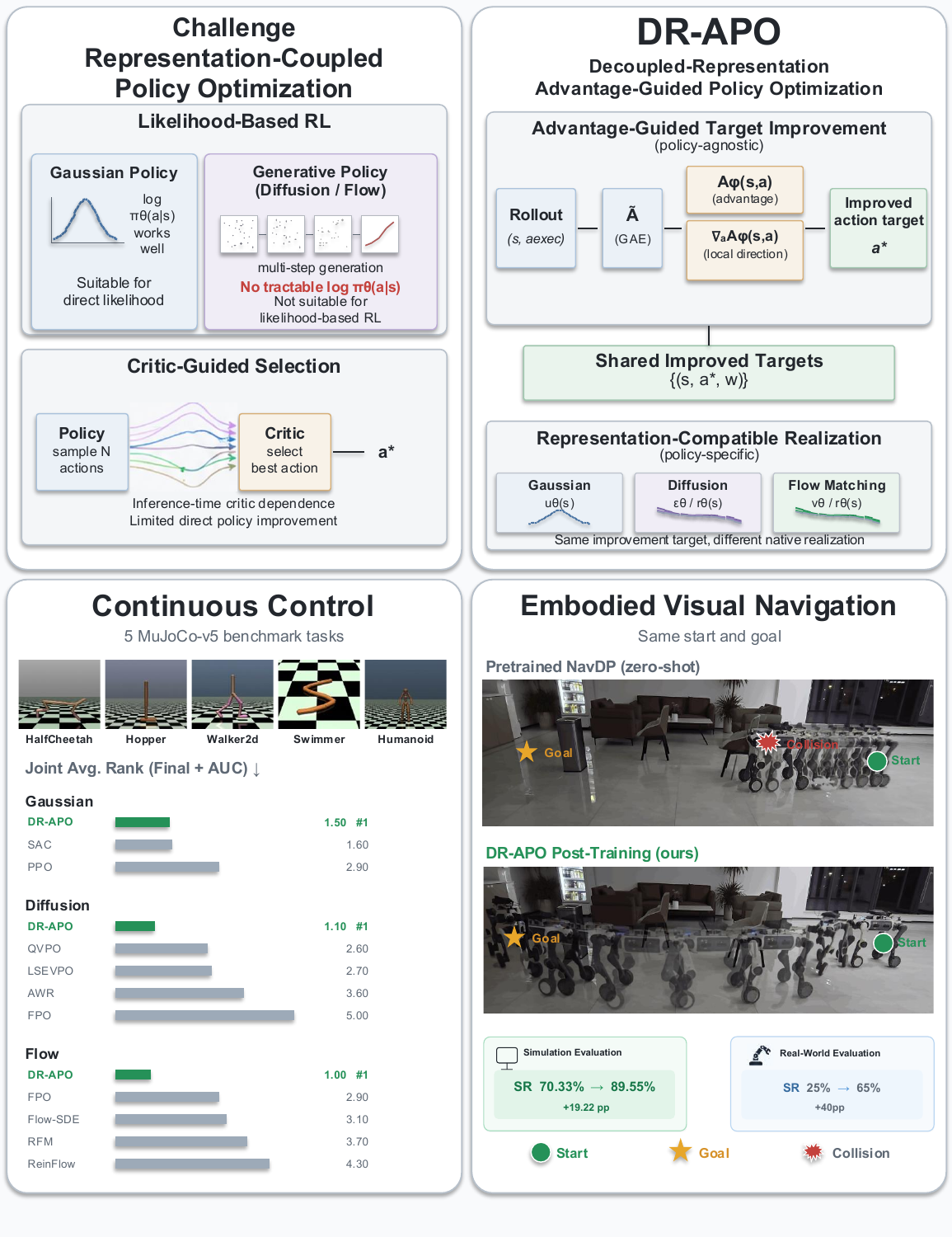}
\caption{Overview of DR-APO. Advantage-guided target improvement and
representation-compatible direct action fitting enable unified optimization
of Gaussian, diffusion, and flow policies across continuous control and
embodied visual navigation.}
\label{fig:atpo_overview}
\vspace{-8mm}
\end{figure}

\section{Introduction}

Mainstream RL policy-optimization methods are largely developed for
specific policy representations, most notably Gaussian policies.
Recent advances in diffusion and flow matching
~\cite{chi2024diffusionpolicy,lipman2023flowmatching}
have enabled expressive generative policies, creating a growing need for
RL methods that generalize across heterogeneous policy representations.
However, existing optimization interfaces remain representation-specific
and transfer poorly to multi-step generative policies.

One line of work extends likelihood-based optimization to generative policies.
While PPO-style likelihood ratios are straightforward for Gaussian policies
~\cite{schulman2017ppo}, diffusion and flow policies require optimization over
their internal generation processes
~\cite{ren2024dppo,ma2025dpmd,zhang2026reinflow,chen2026pirl},
coupling policy updates to multi-step generation and likelihood-ratio
aggregation.

Another line exploits generative sampling to produce multiple candidates for
critic-based evaluation and selection
~\cite{ding2024qvpo,lsevpo,rfm}.
Although effective, execution performance can depend on inference-time critic
selection.
These limitations motivate a unified policy-optimization framework that
improves the underlying policy across representations without explicit
policy-likelihood evaluation or inference-time multi-candidate selection.

We introduce \textbf{DR-APO (Decoupled-Representation Advantage-Guided Policy Optimization)},
a unified framework that decouples policy improvement from the underlying policy
representation.
DR-APO consists of \emph{representation-decoupled policy optimization},
\emph{advantage-guided target improvement}, and
\emph{representation-compatible direct action fitting}.
It first reformulates policy improvement as target optimization independent of
policy parameterization, then constructs locally improved targets from advantage
information, and finally realizes these targets through representation-compatible
fitting.
This enables a unified policy-improvement principle across Gaussian, diffusion,
and flow policies without relying on likelihood-ratio actor updates or
inference-time multi-candidate critic selection.

We evaluate DR-APO on continuous control and closed-loop embodied navigation.
Across five MuJoCo-v5 tasks, DR-APO remains competitive with Gaussian policies,
achieves the best result in $9/10$ diffusion comparisons, and ranks first in
all $10/10$ flow comparisons, without inference-time multi-candidate critic
selection.
DR-APO further improves pretrained NavDP navigation success from $70.33\%$ to
$89.55\%$ in simulation and from $25\%$ to $65\%$ on a DEEP Robotics M20
quadruped robot, a real-world gain of $40$ percentage points.

Our contributions are threefold:
\begin{itemize}
    \item We propose a representation-decoupled policy-optimization framework.
    \item We introduce advantage-guided target improvement and
    representation-compatible direct action fitting across Gaussian, diffusion,
    and flow policies.
    \item We validate DR-APO from MuJoCo continuous control to embodied-policy
    post-training.
\end{itemize}

\section{Related Work}

\subsection{Conventional Policy Optimization}

Continuous-control policy optimization has been extensively studied with
parametric actors.
TRPO and PPO optimize stochastic policies through trust-region or clipped
likelihood-ratio objectives~\cite{schulman2015trpo,schulman2017ppo},
whereas DDPG and TD3 update deterministic actors through learned critic
gradients~\cite{lillicrap2016ddpg,fujimoto2018td3}.
SAC combines off-policy value learning with maximum-entropy optimization
~\cite{haarnoja2018sac}, while MPO and AWR perform non-parametric improvement,
projection, or advantage-weighted regression
~\cite{abdolmaleki2018mpo,peng2019awr}.
Related weighted-fitting methods include RWR, AWAC, and CRR
~\cite{peters2007rwr,nair2020awac,wang2020crr}.
These paradigms are effective for conventional policies, but are generally
developed around specific policy representations.

\subsection{Generative Policy Optimization}

Diffusion and flow models provide expressive multimodal policies for actions
and trajectories~\cite{janner2022diffuser,ajay2023decisiondiffuser,
wang2023diffusionql,chi2024diffusionpolicy,lipman2023flowmatching}.
Their iterative generation processes, however, make conventional optimization
interfaces difficult to transfer directly.
Existing methods mainly address this mismatch through internal-process policy
optimization, value-guided learning or selection, and critic-driven action
refinement.

One line adapts policy-gradient optimization to the generative process.
DPPO treats diffusion denoising as an auxiliary decision process
~\cite{ren2024dppo}, while ReinFlow and $\pi_{\mathrm{RL}}$ introduce
stochastic formulations for flow policies
~\cite{zhang2026reinflow,chen2026pirl}.
FPO develops a PPO-style optimization interface for flow matching
~\cite{mcallister2025fpo}, while GenPO and ADPO further study on-policy or
adaptive optimization of diffusion policies
~\cite{ding2026genpo,jiang2025adpo}.
These methods remain coupled to representation-specific denoising, flow
dynamics, or surrogate objectives.

Another line incorporates learned value information into policy learning or
execution.
QVPO and RFM combine generative policies with critic-guided candidate selection
~\cite{ding2024qvpo,rfm}, while other approaches inject critic information
through score matching, reverse diffusion, or value-reweighted objectives
~\cite{psenka2025qsm,wang2024dacer,ma2025dpmd}.
When multiple candidates are evaluated by a critic, execution performance
additionally depends on inference-time selection.

Critic information can also refine actions before policy fitting.
DIPO applies multi-step $Q$-gradient ascent before retraining a diffusion policy
~\cite{yang2023dipo}, while PA-RL combines critic-based reranking, local action
optimization, and policy distillation~\cite{mark2024parl}.
Related flow-based methods transfer critic guidance into native vector-field
objectives~\cite{shi2026flowdpg}.
Repeated critic-gradient refinement can induce substantial target shifts,
making the resulting supervision sensitive to critic estimation errors.
Overall, existing approaches remain tied to representation-specific
optimization, inference-time critic selection, or critic-driven action
refinement.
A shared policy-improvement mechanism across heterogeneous policy
representations remains less explored.

\subsection{Embodied Generative Policies and Post-Training}

Generative policies are increasingly used in embodied control and navigation.
Diffusion Policy applies diffusion models to visuomotor control
~\cite{chi2024diffusionpolicy}, while NavDP generates multimodal local
trajectories for visual navigation~\cite{navdp}.
Recent work further improves pretrained navigation policies through interaction,
including reward-guided self-imitation~\cite{zhang2026sidp},
$Q$-reweighted post-training~\cite{yang2026xnavdp},
online imitation~\cite{navol}, and GRPO-based fine-tuning~\cite{navdprl}.
These studies demonstrate the effectiveness of embodied-policy post-training
and motivate evaluating whether a general policy-improvement framework can
transfer from continuous control to pretrained embodied policies.

\section{Method}
\label{sec:method}

\begin{figure*}[!t]
\centering
\includegraphics[width=\textwidth]{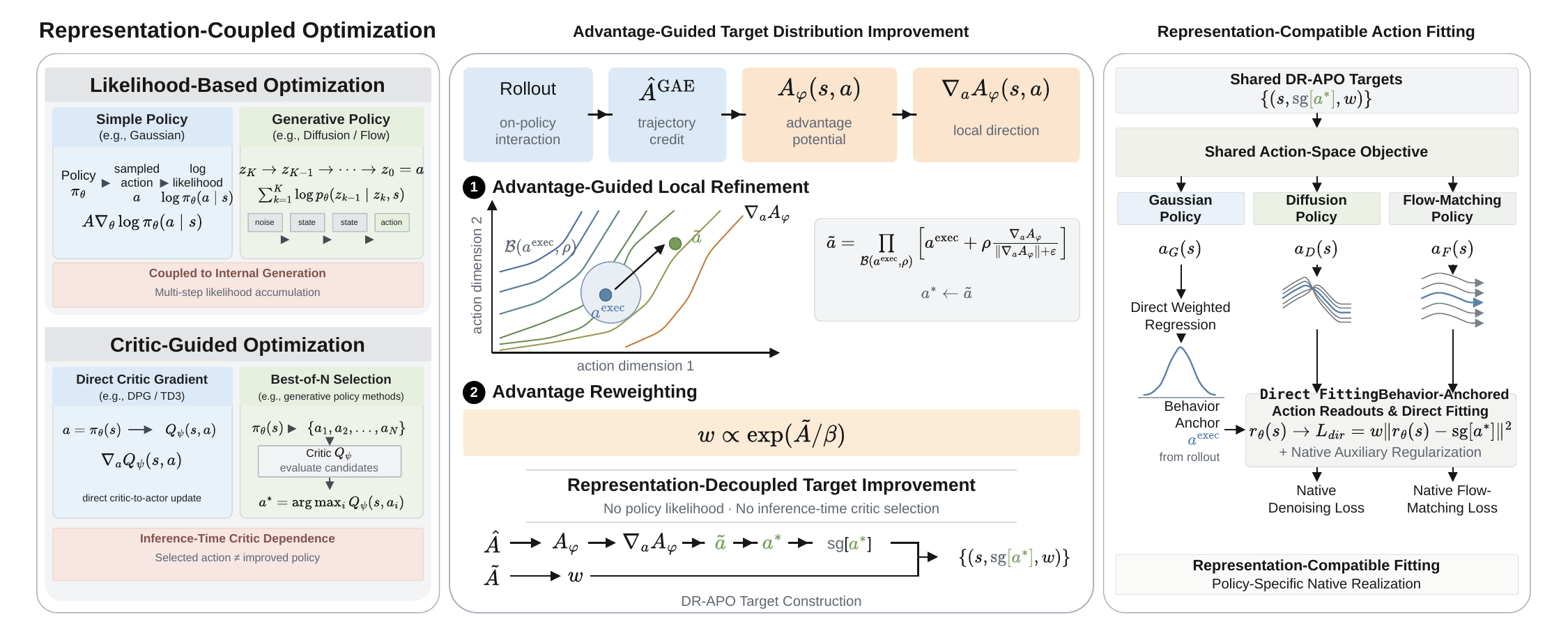}
\caption{Overview of DR-APO. Policy improvement is decoupled from policy
representation through advantage-guided target improvement and
representation-compatible direct action fitting across Gaussian, diffusion,
and flow policies.}
\label{fig:atpo_structure}
\vspace{-4mm}
\end{figure*}

\subsection{Representation-Decoupled Policy Optimization}
\label{sec:decoupled_optimization}

We consider a discounted Markov decision process (MDP) with state $s_t$
and policy output $a_t$.
Reinforcement learning seeks to maximize the expected discounted return:
\begin{equation}
\max_\theta J(\pi_\theta),
\qquad
J(\pi)
=
\mathbb E_\pi
\left[
\sum_{t=0}^{\infty}\gamma^t r_t
\right],
\label{eq:drapo_return}
\end{equation}
where $\theta$ denotes the policy parameters, $r_t$ is the reward received
at decision step $t$, and $\gamma\in(0,1)$ is the discount factor.

Following the performance-difference lemma~\cite{kakade2002approximately}
and the first-order local surrogate used in
TRPO~\cite{schulman2015trpo}, we approximate the policy update at iteration
$k$ as
\begin{equation}
\max_\theta\;
\mathbb E_{s\sim\rho_k,\,
a\sim\pi_\theta(\cdot\mid s)}
\left[
A^{\pi_k}(s,a)
\right],
\label{eq:drapo_surrogate}
\end{equation}
where $\pi_k=\pi_{\theta_k}$ is the current policy used to collect
iteration-$k$ rollouts, $A^{\pi_k}(s,a)$ is its advantage function, and
$\rho_k$ is its normalized discounted state-visitation distribution.

Introducing an auxiliary conditional action distribution $p(\cdot\mid s)$,
we equivalently reformulate~\eqref{eq:drapo_surrogate} as
\begin{equation}
\begin{aligned}
\max_{p,\theta}\quad&
\mathbb E_{s\sim\rho_k,\,
a\sim p(\cdot\mid s)}
\left[
A^{\pi_k}(s,a)
\right]
\\
\mathrm{s.t.}\quad&
p(\cdot\mid s)=\pi_\theta(\cdot\mid s).
\end{aligned}
\label{eq:drapo_constrained_formulation}
\end{equation}
Replacing this hard constraint with a distribution-consistency penalty
yields the relaxed optimization problem
\begin{equation}
\begin{aligned}
\max_{p,\theta}\;\Bigl\{
&
\mathbb E_{s\sim\rho_k,\,
a\sim p(\cdot\mid s)}
\left[
A^{\pi_k}(s,a)
\right]
\\
&-
\lambda_{\rm c}\mathcal D_k(p,\pi_\theta)
\Bigr\},
\end{aligned}
\label{eq:drapo_relaxed_formulation}
\end{equation}
where $\lambda_{\rm c}>0$ is the consistency penalty coefficient and
$\mathcal D_k$ is a nonnegative discrepancy between the two conditional
action distributions, averaged over states drawn from $\rho_k$.

We optimize~\eqref{eq:drapo_relaxed_formulation} by coordinate ascent,
alternating between the target distribution and policy parameters,
analogous to the treatment in relative-entropy policy
optimization~\cite{abdolmaleki2018mpo}.
At iteration $k$, we first fix the current policy
$\pi_k=\pi_{\theta_k}$ and optimize with respect to $p$, obtaining
\begin{equation}
\begin{aligned}
p_k^+
\in
\arg\max_p\;\Bigl\{
&
\mathbb E_{s\sim\rho_k,\,
a\sim p(\cdot\mid s)}
\left[
A^{\pi_k}(s,a)
\right]
\\
&-
\lambda_{\rm c}\mathcal D_k(p,\pi_k)
\Bigr\}.
\end{aligned}
\label{eq:drapo_distribution_improvement}
\end{equation}
Here, $p_k^+$ denotes the improvement target for the current iteration
relative to $\pi_k$, rather than a globally optimal policy distribution.

With $p_k^+$ fixed, the expected-advantage term
in~\eqref{eq:drapo_relaxed_formulation} is independent of $\theta$.
The remaining optimization therefore reduces to minimizing the
distribution-consistency error:
\begin{equation}
\theta_{k+1}
\in
\arg\min_\theta
\mathcal D_k\left(p_k^+,\pi_\theta\right).
\label{eq:drapo_distribution_fitting}
\end{equation}

This decomposition yields two successive stages of policy optimization:

\begin{itemize}
    \item \textbf{Representation-decoupled target improvement:}
    construct an improved target distribution $p_k^+$ from interaction feedback,
    independent of the policy's internal parameterization.

    \item \textbf{Representation-dependent target fitting:}
    fit $p_k^+$ through the learning interface of the selected policy representation.
\end{itemize}

Thus, the same improvement rule can be shared across Gaussian, diffusion, and
flow-matching policies, while only the fitting interface is representation-specific.

\subsection{Advantage-Guided Target Distribution Improvement}
\label{sec:drapo_target_improvement}

Equation~\eqref{eq:drapo_distribution_improvement} provides a general
optimization problem for the target distribution $p_k^+$.
We specify the distribution-consistency penalty as
\begin{equation}
\lambda_{\rm c}\mathcal D_k(p,\pi_k)
=
\mathbb E_{s\sim\rho_k}
\inf_q
\left[
\beta D_{\rm KL}(q\|\pi_k)
+
\frac{1}{2\eta}W_2^2(q,p)
\right],
\label{eq:drapo_consistency_geometry}
\end{equation}
where $p(\cdot\mid s)$ is the target action distribution being optimized,
$q(\cdot\mid s)$ is an auxiliary action distribution, and
$\beta,\eta>0$ control probability redistribution and action displacement,
respectively.
Here, $W_2$ denotes the second-order Wasserstein distance.
Substituting~\eqref{eq:drapo_consistency_geometry}
into~\eqref{eq:drapo_distribution_improvement} gives, for each fixed
state $s$,
\begin{equation}
\begin{aligned}
\max_{p,q}\;\Bigl\{
&\mathbb E_{a'\sim p}[A^{\pi_k}(s,a')]
-\frac{1}{2\eta}W_2^2(q,p)
\\
&-\beta D_{\rm KL}(q\|\pi_k)
\Bigr\},
\end{aligned}
\label{eq:drapo_joint_target}
\end{equation}
where the conditioning on $s$ is omitted from the distributions.

Using the optimal-coupling representation of $W_2$, we first optimize
over $p$, obtaining
\begin{equation}
\begin{aligned}
\Phi_{k,\eta}(s,a)
&\in
\arg\max_{a'\in\mathcal A}
\left[
A^{\pi_k}(s,a')
-\frac{\|a'-a\|_2^2}{2\eta}
\right],
\\
A_{k,\eta}^{+}(s,a)
&=
A^{\pi_k}\!\left(s,\Phi_{k,\eta}(s,a)\right)
\\
&\quad-
\frac{\|\Phi_{k,\eta}(s,a)-a\|_2^2}{2\eta}.
\end{aligned}
\label{eq:drapo_proximal_action}
\end{equation}
Here, $\Phi_{k,\eta}$ is the ideal action-improvement map, and
$A_{k,\eta}^{+}$ is the proximal improvement value after subtracting
the displacement cost.
Subsequent optimization over $q$ yields
\begin{equation}
\begin{aligned}
q_k^+(a\mid s)
&=
\frac{
\pi_k(a\mid s)
\exp\!\left(A_{k,\eta}^{+}(s,a)/\beta\right)
}{
Z_{k,\eta}(s)
},
\\
Z_{k,\eta}(s)
&=
\mathbb E_{a\sim\pi_k(\cdot\mid s)}
\left[
e^{A_{k,\eta}^{+}(s,a)/\beta}
\right],
\end{aligned}
\label{eq:drapo_auxiliary_target}
\end{equation}
where $Z_{k,\eta}(s)$ is the normalizing factor.
The resulting target distribution is
\begin{equation}
\begin{aligned}
p_k^+(\mathrm da'\mid s)
&=
\int
\delta_{\Phi_{k,\eta}(s,a)}(\mathrm da')
\\
&\qquad\times
\frac{e^{A_{k,\eta}^{+}(s,a)/\beta}}{Z_{k,\eta}(s)}
\,\pi_k(\mathrm da\mid s),
\end{aligned}
\label{eq:drapo_target_pushforward}
\end{equation}
where $\delta_x$ denotes the Dirac measure at $x$.
Equation~\eqref{eq:drapo_target_pushforward} gives the ideal
action-then-weight construction: first determine the improved action
locations, then assign probability mass according to their proximal
improvement values.

\textit{Finite-Sample Policy Improvement.}
A common class of policy-improvement methods approximates the target
distribution by sampling $K$ candidate actions at the same state and
exponentially reweighting them according to their advantages.
Let $a_j\overset{\mathrm{i.i.d.}}{\sim}\pi_k(\cdot\mid s)$ and
$A_j=A^{\pi_k}(s,a_j)$. Define
\begin{equation}
\begin{aligned}
\omega_j
&=
\frac{\exp(A_j/\beta)}
{\sum_{\ell=1}^{K}\exp(A_\ell/\beta)},
\\
\Gamma_K^{\rm multi}(s)
&:=
\mathbb E
\left[
\sum_{j=1}^{K}\omega_jA_j
-\frac1K\sum_{j=1}^{K}A_j
\right].
\end{aligned}
\label{eq:drapo_multiaction_gain}
\end{equation}
Here, $\omega_j$ is the normalized weight of candidate $j$, and
$\Gamma_K^{\rm multi}(s)$ is the expected advantage gain over uniform
candidate weighting, with expectation over the $K$ candidate draws.
Let
$\sigma_A^2(s)
:=
\operatorname{Var}_{a\sim\pi_k(\cdot\mid s)}
[A^{\pi_k}(s,a)]$
denote the action-wise advantage variance.
Under weak exponential tilting,
$|A^{\pi_k}(s,a)|/\beta\ll1$,
\begin{equation}
\Gamma_K^{\rm multi}(s)
\simeq
\frac{K-1}{K\beta}\sigma_A^2(s).
\label{eq:drapo_weak_tilt_gain}
\end{equation}

DR-APO instead uses a single executed action
$a\sim\pi_k(\cdot\mid s)$ and refines it using first-order advantage
information. For a nonzero gradient, the idealized update and its
expected advantage gain are
\begin{equation}
\begin{aligned}
a^+
&=
a+
\rho
\frac{\nabla_aA^{\pi_k}(s,a)}
{\|\nabla_aA^{\pi_k}(s,a)\|_2},
\\
\Gamma_1^{\rm ref}(s)
&:=
\mathbb E_{a\sim\pi_k(\cdot\mid s)}
\left[
A^{\pi_k}(s,a^+)-A^{\pi_k}(s,a)
\right].
\end{aligned}
\label{eq:drapo_single_action_gain}
\end{equation}
Here, $a^+$ is the refined action, $\rho>0$ is the refinement radius,
and $\nabla_aA^{\pi_k}$ is the action gradient of the advantage.
If $A^{\pi_k}(s,\cdot)$ has an $L_A$-Lipschitz gradient and the update
remains feasible, then
\[
\begin{aligned}
\Gamma_1^{\rm ref}(s)
\ge{}&
\rho\,
\mathbb E_{a\sim\pi_k(\cdot\mid s)}
\left[
\|\nabla_aA^{\pi_k}(s,a)\|_2
\right]
-\frac{L_A\rho^2}{2}.
\end{aligned}
\]
Consequently, within the approximation
in~\eqref{eq:drapo_weak_tilt_gain}, the condition
\begin{equation}
\begin{aligned}
&\mathbb E_{a\sim\pi_k(\cdot\mid s)}
\left[
\|\nabla_aA^{\pi_k}(s,a)\|_2
\right]
\\
&\qquad\ge
\frac{K-1}{K\beta\rho}\sigma_A^2(s)
+\frac{L_A\rho}{2}
\end{aligned}
\label{eq:drapo_gradient_threshold}
\end{equation}
gives
\[
\Gamma_1^{\rm ref}(s)
\ge
\Gamma_K^{\rm multi}(s).
\]
Thus, under this approximation, a single executed action can provide
an expected refinement gain no smaller than that of
$K$-candidate exponential reweighting when its average
advantage-gradient magnitude satisfies
\eqref{eq:drapo_gradient_threshold}.
This idealized comparison concerns exponential reweighting, not
inference-time Best-of-$N$ selection.

\textit{Practical Realization.}
Given a rollout batch
$\mathcal B_k=\{(s_i,a_i^{\rm exec},\hat A_i)\}_{i=1}^{B}$,
we fit a differentiable advantage estimator $A_\phi$ to the rollout
advantage estimates $\hat A_i$ and compute its action gradient at each
executed action:
\begin{equation}
\begin{aligned}
\mathcal L_A(\phi)
&=
\frac1B\sum_{i=1}^{B}
\left[
A_\phi(s_i,a_i^{\rm exec})
-\operatorname{sg}(\hat A_i)
\right]^2,
\\
g_i
&=
\left.
\nabla_aA_\phi(s_i,a)
\right|_{a=a_i^{\rm exec}},
\end{aligned}
\label{eq:drapo_advantage_fit}
\end{equation}
where $\operatorname{sg}$ denotes stop-gradient.
The estimator $A_\phi$ supplies local action directions, while
rollout advantages determine sample weights.
We then construct a bounded refinement:
\begin{equation}
\begin{aligned}
a_i^+
&=
\Pi_{\mathcal C_i}
\left[
a_i^{\rm exec}
+
\rho\frac{g_i}{\|g_i\|_2+\varepsilon}
\right],
\\
\mathcal C_i
&=
\mathcal A\cap\mathcal B_2(a_i^{\rm exec},\rho).
\end{aligned}
\label{eq:drapo_bounded_refinement}
\end{equation}
Here, $\mathcal B_2(a_i^{\rm exec},\rho)$ is the Euclidean ball of
radius $\rho$, $\Pi_{\mathcal C_i}$ denotes Euclidean projection,
and $\varepsilon>0$ provides numerical stability.
The resulting displacement $\delta_i:=a_i^+-a_i^{\rm exec}$
satisfies $\|\delta_i\|_2\le\rho$ by construction.

In practice, we use the rollout advantage grounded in the executed
behavior as a reweighting proxy, rather than explicitly evaluating
the proximal improvement value.
With standardized advantages $\tilde A_i$, we obtain
\begin{equation}
\begin{aligned}
w_i
&=
\frac1C
\exp\!\left(
\frac{\tilde A_i}{\beta_{\rm w}}
\right),
\\
\mathcal T_k^+
&=
\{(s_i,a_i^+,w_i)\}_{i=1}^{B},
\end{aligned}
\label{eq:drapo_weighted_targets}
\end{equation}
where $\beta_{\rm w}>0$ is the weighting temperature and $C>0$ is a
common scaling constant.
These weights specify relative fitting importance rather than
per-state normalized probabilities.
Equations~\eqref{eq:drapo_advantage_fit}--\eqref{eq:drapo_weighted_targets}
provide a finite-sample realization of the action-then-weight
structure in~\eqref{eq:drapo_target_pushforward}, rather than an
exact solution of the ideal distributional update.
Different policy representations subsequently fit the same target set
$\mathcal T_k^+$.

\subsection{Representation-Compatible Direct Action Fitting}
\label{sec:direct_action_fitting}

From Sec.~\ref{sec:drapo_target_improvement}, we obtain
\begin{equation}
\mathcal T_k^+=\{(s_i,a_i^+,w_i)\}_{i=1}^{B},
\quad a_i^+=a_i^{\rm exec}+\delta_i,
\quad \|\delta_i\|_2\le\rho .
\tag{19}
\end{equation}
The fitting stage seeks to realize these local refinements within a finite
policy update. For Gaussian policies, we directly fit the policy mean to
the refined action target, defining the action readout as
\begin{equation}
r_\theta^G(s_i)=\mu_\theta(s_i).
\tag{20}
\end{equation}
This readout is the distribution mean, rather than a sampled action.

Generative policies instead predict a diffusion noise field~\cite{ho2020ddpm}
or flow velocity field. A native-fitting baseline uses $a_i^+$ as the clean target:
\begin{equation}
\begin{aligned}
\ell_{\rm nat}^{D}
&=\frac12\left\|
\epsilon_\theta(s_i,\alpha_\tau a_i^++\sigma_\tau\epsilon,\tau)
-\epsilon\right\|_2^2,\\
\ell_{\rm nat}^{F}
&=\frac12\left\|
v_\theta(s_i,(1-\tau)a_i^++\tau\epsilon,\tau)
-(\epsilon-a_i^+)\right\|_2^2.
\end{aligned}
\tag{21}
\end{equation}
Although native fitting can learn the refined targets, it first translates
$\delta_i$ into a representation-specific regression perturbation.
Let $z_\tau^{\mathcal R}$ and $y_\tau^{\mathcal R}$ denote its input and target:
\begin{equation}
\begin{aligned}
\Delta z_{i,\tau}^{\mathcal R}
&=z_\tau^{\mathcal R}(a_i^+,\epsilon)
-z_\tau^{\mathcal R}(a_i^{\rm exec},\epsilon),\\
\Delta y_{i,\tau}^{\mathcal R}
&=y_\tau^{\mathcal R}(a_i^+,\epsilon)
-y_\tau^{\mathcal R}(a_i^{\rm exec},\epsilon).
\end{aligned}
\tag{22}
\end{equation}
For small refinements and negligible native residuals at the behavior
inputs, the first-order change in the native-loss gradient is
\begin{equation}
\Delta g_i^{\mathcal R}\approx
(J_\theta^{\mathcal R})^\top
\left(J_z^{\mathcal R}\Delta z_{i,\tau}^{\mathcal R}
-\Delta y_{i,\tau}^{\mathcal R}\right),
\tag{23}
\end{equation}
where $J_\theta^{\mathcal R}$ and $J_z^{\mathcal R}$ are predictor Jacobians
evaluated at the current policy and behavior input.
Specifically,
\begin{equation}
\begin{aligned}
\Delta z_{i,\tau}^{D}&=\alpha_\tau\delta_i,
&\Delta y_{i,\tau}^{D}&=0,\\
\Delta z_{i,\tau}^{F}&=(1-\tau)\delta_i,
&\Delta y_{i,\tau}^{F}&=-\delta_i,
\end{aligned}
\tag{24}
\end{equation}
which gives
\begin{equation}
\begin{aligned}
\Delta g_i^{D}
&\approx\alpha_\tau(J_\theta^\epsilon)^\top J_z^\epsilon\delta_i,\\
\Delta g_i^{F}
&\approx(J_\theta^v)^\top\left[(1-\tau)J_z^v+I\right]\delta_i.
\end{aligned}
\tag{25}
\end{equation}
Thus, the refinement enters through field-level changes; the diffusion
input perturbation is additionally scaled by $\alpha_\tau$.

A local field correction also propagates through the generator.
For $x_{j-1}=F_j^\theta(x_j,f_j)$ and $a=x_0$, an isolated prediction
change $\Delta f_j$ has first-order endpoint effect
\begin{equation}
\Delta a^{(j)}\approx
\left(\prod_{\ell=1}^{j-1}
\frac{\partial F_\ell}{\partial x_\ell}\right)
\frac{\partial F_j}{\partial f_j}\Delta f_j.
\tag{26}
\end{equation}
For $K$-step Euler flow with step size $1/K$ and state-transition Jacobian
products bounded by $C$,
\begin{equation}
\|\Delta a^{(j)}\|_2\le\frac{C}{K}\|\Delta v_j\|_2.
\tag{27}
\end{equation}
This describes a single-step perturbation, not the complete parameter
update, which can affect multiple generation steps.

\textit{Behavior-Anchored Action Readouts.}
For generative policies, we construct differentiable clean-action
estimates at behavior-anchored inputs. Auxiliary noise $\epsilon_{i,m}$
is introduced specifically for fitting:
\begin{equation}
\begin{aligned}
z_{i,\tau_m}^{D}
&=\alpha_{\tau_m}a_i^{\rm exec}+\sigma_{\tau_m}\epsilon_{i,m},\\
z_{i,\tau_m}^{F}
&=(1-\tau_m)a_i^{\rm exec}+\tau_m\epsilon_{i,m}.
\end{aligned}
\tag{28}
\end{equation}
The corresponding diffusion and flow readouts are
\begin{equation}
r_\theta^D(s,z,\tau)
=\frac{z-\sigma_\tau\epsilon_\theta(s,z,\tau)}{\alpha_\tau},
\tag{29}
\end{equation}
\begin{equation}
r_\theta^F(s,z,\tau)=z-\tau v_\theta(s,z,\tau).
\tag{30}
\end{equation}
These follow from the native corruption or interpolation equations,
recovering the clean action for an exact samplewise noise or velocity
prediction. Such samplewise recovery is not assumed for learned predictors.
We average $M$ intermediate readouts:
\begin{equation}
r_\theta^{\mathcal R}(s_i,\xi_i)
=\frac1M\sum_{m=1}^{M}
r_\theta^{\mathcal R}(s_i,z_{i,\tau_m}^{\mathcal R},\tau_m),
\quad\mathcal R\in\{D,F\},
\tag{31}
\end{equation}
where $\xi_i=\{(z_{i,\tau_m}^{\mathcal R},\tau_m)\}_{m=1}^{M}$
is fixed during the corresponding policy update. These are inputs
constructed from the behavior action, not states from its rollout
sampling path. Unlike Eq.~(21), the direct objective keeps these inputs
anchored at $a_i^{\rm exec}$ while supervising the readouts with $a_i^+$.

\textit{Direct Action Fitting.}
All three policy families share the weighted action-space objective
\begin{equation}
\mathcal L_{\rm dir}^{\mathcal R}
=\frac1B\sum_{i=1}^{B}\frac{w_i}{2}
\left\|r_\theta^{\mathcal R}(s_i,\xi_i)
-\operatorname{sg}(a_i^+)\right\|_2^2,
\tag{32}
\end{equation}
where $r_\theta^G(s_i,\xi_i):=\mu_\theta(s_i)$ is independent of
$\xi_i$ for Gaussian policies.
Define $e_i^{\mathcal R}=r_{\theta_k}^{\mathcal R}(s_i,\xi_i)
-a_i^{\rm exec}$ and
$J_i^r=\left.\partial r_\theta^{\mathcal R}(s_i,\xi_i)/\partial\theta
\right|_{\theta_k}$.
The unweighted per-sample gradient is exactly
\begin{equation}
h_i^{\mathcal R}
:=\left.\nabla_\theta\ell_{\rm dir}^{\mathcal R}\right|_{\theta_k}
=(J_i^r)^\top(e_i^{\mathcal R}-\delta_i).
\tag{33}
\end{equation}
Thus, $\delta_i$ enters the action-space residual explicitly.
For Gaussian policies, $e_i^G=\mu_{\theta_k}(s_i)-a_i^{\rm exec}$
is the mean-to-behavior-action offset; for generative policies,
$e_i^{\mathcal R}$ is the behavior-readout reconstruction residual.
Neither is assumed to vanish.
For a smooth loss and sufficiently small $\eta_\theta$,
\begin{equation}
\begin{aligned}
\ell_{\rm dir}^{\mathcal R}(\theta_k-\eta_\theta h_i^{\mathcal R})
&=\ell_{\rm dir}^{\mathcal R}(\theta_k)\\
&\quad-\eta_\theta\|h_i^{\mathcal R}\|_2^2
+\mathcal O(\eta_\theta^2).
\end{aligned}
\tag{34}
\end{equation}
This local descent concerns the single-sample readout loss, not a guarantee
of final-action displacement or return improvement. The weights $w_i$
scale each target's contribution to the shared objective.

Since selected readouts do not constrain the complete generative field,
we retain native training as an auxiliary regularizer:
\begin{equation}
\mathcal L^{\mathcal R}
=\mathcal L_{\rm dir}^{\mathcal R}
+\lambda_{\rm nat}\mathcal L_{\rm aux}^{\mathcal R},
\qquad\mathcal R\in\{D,F\}.
\tag{35}
\end{equation}
The direct term supervises local action refinements, while the auxiliary
term preserves the corresponding native field-learning signal. Gaussian
policies use mean regression under the direct objective in Eq.~(32).

\section{Experiments}
\label{sec:experiments}
\begin{figure*}[t]
\centering
\makebox[\textwidth][c]{\includegraphics[width=0.8\textwidth,trim=29.6pt 16pt 4.7pt 0pt,clip]{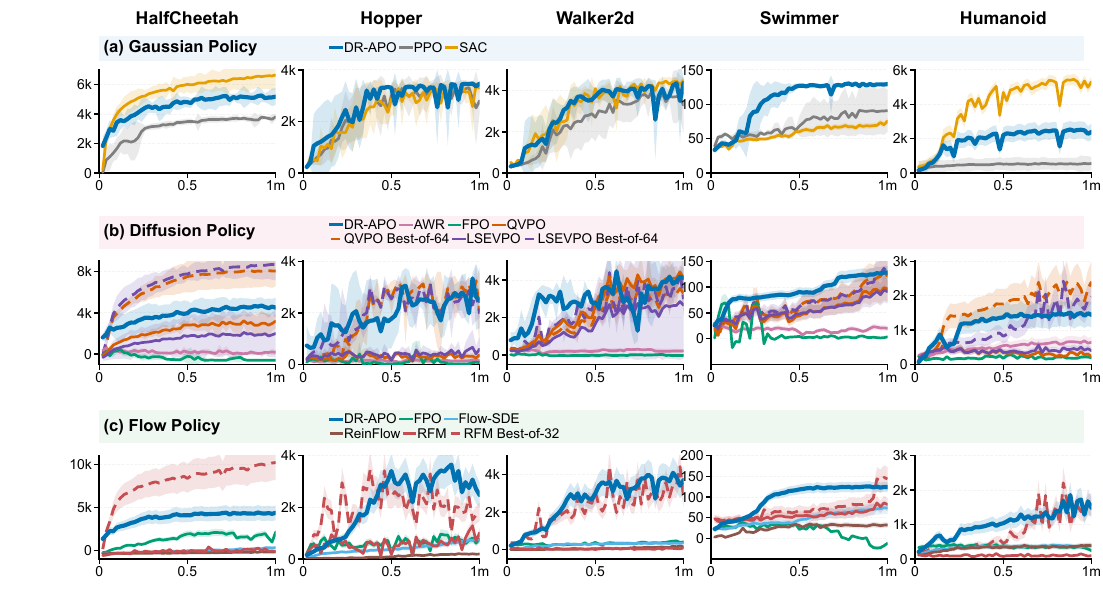}}
\caption{Learning curves of Gaussian, diffusion, and flow policies on five
MuJoCo-v5 tasks. Curves and shaded regions denote the mean and standard deviation.}
\label{fig:learning_curve}
\vspace{-4mm}
\end{figure*}

We evaluate DR-APO across Gaussian, diffusion, and flow policies on five
MuJoCo-v5 tasks, ablate its key components, and assess NavDP post-training
in simulated and real-world navigation.

\subsection{Experimental Setup}
\label{sec:experimental-setup}

\paragraph{Continuous control}
We train policies from scratch for $1$M steps on five MuJoCo-v5
environments~\cite{todorov2012mujoco}:
HalfCheetah (HC), Hopper (Hop), Walker2d (W2), Swimmer (Swi), and Humanoid (Hum).
Policies are evaluated every $20$k steps over five random seeds.
Gaussian policies are compared with PPO and SAC; diffusion policies with AWR,
FPO, QVPO, and LSEVPO; and flow policies with FPO, Flow-SDE, ReinFlow, and RFM.
Best-of-64 variants of QVPO/LSEVPO and the Best-of-32 variant of RFM require
additional critic-based selection, and are therefore reported separately and
excluded from direct-policy rankings.
We report \textbf{Final Return} over $820$k--$1$M steps, normalized
\textbf{AUC@1M} over $20$k--$1$M steps, and the average rank across the five
environments.

\paragraph{Embodied navigation}
We post-train a pretrained diffusion-based NavDP policy with DR-APO using
a wheeled robot in InternScenes~\cite{internscenes}. Both simulation training
and testing use this wheeled platform, with an unchanged execution controller;
we report \textbf{SR} and \textbf{SPL}.
Real-world trials use a DEEP Robotics M20 quadruped robot to compare
pretrained NavDP with DR-APO post-training using SR
(Table~\ref{tab:real_navigation}).

\subsection{Cross-Representation Policy Learning}
\label{sec:cross-representation}

Fig.~\ref{fig:learning_curve} and Table~\ref{tab:main_results} summarize the results across five MuJoCo-v5 environments.
Under Gaussian policies, DR-APO achieves the best average AUC@1M rank while
remaining competitive with SAC in Final Return.
For diffusion policies, DR-APO achieves the highest mean AUC@1M
on all five tasks and the highest mean Final Return on four,
leading in 9 of the 10 environment--metric comparisons.
For flow policies, DR-APO ranks first on both metrics across
all five tasks.

These results show that the same target-level policy-improvement rule transfers
across Gaussian, diffusion, and flow representations without requiring
representation-specific likelihood optimization over their internal generation
processes.
Moreover, all DR-APO results are obtained by direct policy execution, without
inference-time multi-candidate critic selection.
Together, the results support representation-decoupled policy improvement while
retaining strong within-family performance for both generative policy classes.

\begin{table*}[t]
\centering
\caption{
Policy-only results on five MuJoCo-v5 tasks. We report Final Return,
AUC@1M, and within-family average ranks. 
}
\label{tab:main_results}
\begingroup
\fontsize{6.5}{6.7}\selectfont
\setlength{\tabcolsep}{1.25pt}
\renewcommand{\arraystretch}{0.88}
\begin{tabular*}{\textwidth}{
@{\extracolsep{\fill}}
ll
*{5}{cc}
@{\hspace{1.25pt}}
cc
@{}
}
\toprule
\multirow{2}{*}{Policy}
& \multirow{2}{*}{Method}
& \multicolumn{2}{c}{HalfCheetah}
& \multicolumn{2}{c}{Hopper}
& \multicolumn{2}{c}{Walker2d}
& \multicolumn{2}{c}{Swimmer}
& \multicolumn{2}{c}{Humanoid}
& \multicolumn{2}{c}{Avg. rank $\downarrow$}
\\
\cmidrule(lr){3-4}
\cmidrule(lr){5-6}
\cmidrule(lr){7-8}
\cmidrule(lr){9-10}
\cmidrule(lr){11-12} 
\cmidrule(l){13-14}
&
& \textbf{Final Return} & \textbf{AUC@1M}
& \textbf{Final Return} & \textbf{AUC@1M}
& \textbf{Final Return} & \textbf{AUC@1M}
& \textbf{Final Return} & \textbf{AUC@1M}
& \textbf{Final Return} & \textbf{AUC@1M}
& Final & AUC
\\

\midrule

\multirow[c]{3}{*}{Gaussian}
& PPO
& 3250.1 & 3057.5
& 3022.9 & 2484.4
& 3652.1 & 2547.9
& 68.4   & 53.4
& 541.0  & 493.2
& 3.00   & 2.80
\\
& SAC
& \textbf{6538.1} & \textbf{5621.8}
& 3158.3 & 2420.7
& \textbf{4343.8} & 3181.8
& 69.8   & 58.0
& \textbf{5137.5} & \textbf{3951.5}
& \textbf{1.40} & 1.80
\\
& \textbf{DR-APO}
& 5154.5 & 4574.1
& \textbf{3288.9} & \textbf{2692.3}
& 4243.2 & \textbf{3202.6}
& \textbf{128.7} & \textbf{106.5}
& 2363.9 & 1945.0
& 1.60 & \textbf{1.40}
\\

\midrule

\multirow[c]{5}{*}{Diffusion}
& AWR
& 135.5 & 205.7
& 144.9 & 143.9
& 220.2 & 228.1
& 19.2  & 17.3
& 640.6 & 509.2
& 3.60  & 3.60
\\
& FPO
& -587.5 & -344.9
& 40.7   & 119.4
& -2.0   & -0.2
& 2.8    & 11.1
& 201.9  & 197.1
& 5.00   & 5.00
\\
& QVPO
& 2981.3 & 2350.8
& 324.4  & 291.4
& \textbf{3913.3} & 2519.4
& 88.9   & 54.7
& 292.1  & 340.5
& 2.40   & 2.80
\\
& LSEVPO
& 1924.6 & 1462.1
& 497.8  & 396.9
& 2521.9 & 1627.0
& 83.0   & 54.9
& 439.4  & 395.8
& 2.80   & 2.60
\\
& \textbf{DR-APO}
& \textbf{4572.7} & \textbf{3840.5}
& \textbf{2631.5} & \textbf{1955.6}
& 3853.6 & \textbf{2902.6}
& \textbf{124.9} & \textbf{93.7}
& \textbf{1462.0} & \textbf{1185.0}
& \textbf{1.20} & \textbf{1.00}
\\

\midrule

\multirow[c]{5}{*}{Flow}
& FPO
& 1601.1 & 1311.8
& 736.2  & 619.0
& 390.5  & 314.3
& -11.2  & 18.4
& 323.9  & 359.3
& 3.20   & 2.60
\\
& Flow-SDE
& 215.7  & -65.0
& 677.9  & 405.2
& 334.3  & 263.0
& 70.1   & 47.5
& 375.3  & 334.8
& 3.00   & 3.20
\\
& ReinFlow
& -211.8 & -295.6
& 185.4  & 97.5
& 135.0  & 52.9
& 31.1   & 24.9
& 367.3  & 316.4
& 4.20   & 4.40
\\
& RFM
& 46.0   & -117.8
& 806.1  & 584.5
& 48.4   & 40.5
& 72.4   & 55.2
& 104.1  & 98.3
& 3.60   & 3.80
\\
& \textbf{DR-APO}
& \textbf{4299.2} & \textbf{3824.2}
& \textbf{3016.4} & \textbf{2212.7}
& \textbf{3650.0} & \textbf{2624.2}
& \textbf{124.2}  & \textbf{98.6}
& \textbf{1535.4} & \textbf{1037.6}
& \textbf{1.00}   & \textbf{1.00}
\\
\bottomrule
\end{tabular*}
\endgroup
\vspace{-15pt}
\end{table*}

\subsection{Ablation and Analysis}
\label{sec:ablation}

\paragraph{Comparison with Best-of-$N$}

\begin{table}[!t]
\centering
\caption{Direct policy execution versus Best-of-$N$ critic selection.}
\label{tab:selection}

\begingroup
\fontsize{6.5}{6.7}\selectfont
\setlength{\tabcolsep}{1.1pt}
\renewcommand{\arraystretch}{0.95}
\setlength{\aboverulesep}{0.15ex}
\setlength{\belowrulesep}{0.15ex}
\newcommand{\sel}[2]{#1$\!\rightarrow\!$#2}

\begin{tabular*}{\columnwidth}{
@{\extracolsep{\fill}}
lccc@{\hspace{2pt}}cc@{}
}
\toprule
& \multicolumn{3}{c}{\textit{Diffusion Policy}}
& \multicolumn{2}{c}{\textit{Flow Policy}} \\
\cmidrule(lr){2-4}
\cmidrule(l){5-6}

Env.
& \shortstack{QVPO\\[-1.5pt]{\scriptsize P$\rightarrow$B64}}
& \shortstack{LSEVPO\\[-1.5pt]{\scriptsize P$\rightarrow$B64}}
& \shortstack{\textbf{DR-APO}\\[-1.5pt]{\scriptsize Direct}}
& \shortstack{RFM\\[-1.5pt]{\scriptsize P$\rightarrow$B32}}
& \shortstack{\textbf{DR-APO}\\[-1.5pt]{\scriptsize Direct}}
\\
\midrule

HC  & \sel{2981.3}{8047.0} & \sel{1924.6}{8591.7} & 4572.7
    & \sel{46.0}{9862.7}    & 4299.2 \\
Hop & \sel{324.4}{2792.6}  & \sel{497.8}{2506.8}  & 2631.5
    & \sel{806.1}{1839.8}   & 3016.4 \\
W2  & \sel{3913.3}{3130.7} & \sel{2521.9}{3316.4} & 3853.6
    & \sel{48.4}{3211.1}    & 3650.0 \\
Swi & \sel{88.9}{113.3}    & \sel{83.0}{112.1}    & 124.9
    & \sel{72.4}{109.4}     & 124.2 \\
Hum & \sel{292.1}{2191.2}  & \sel{439.4}{1797.9}  & 1462.0
    & \sel{104.1}{1470.7}   & 1535.4 \\

\bottomrule
\end{tabular*}

\vspace{3pt}
\scriptsize
P: direct policy execution; B64/B32: critic-selected Best-of-64/32.
\vspace{-8pt}
\endgroup
\end{table}

As shown in Table II, within the corresponding policy families,
DR-APO improves direct-policy Final Return over LSEVPO and RFM
on all five tasks, and over QVPO on four, without inference-time
critic selection.
For flow policies, DR-APO further outperforms RFM Best-of-32 on four of the
five tasks, with HalfCheetah being the exception.
For diffusion policies, the comparison is task dependent: DR-APO surpasses
LSEVPO Best-of-64 on Hopper, Walker2d, and Swimmer, while trailing it on
HalfCheetah and Humanoid.

\paragraph{Direct Action Fitting}
\begin{table}[!t]
\centering
\caption{
Ablation studies on target improvement and direct action fitting.
Final Return is reported as mean $\pm$ SD over five seeds.
}
\label{tab:ablation}

\begingroup
\fontsize{6.15}{6.45}\selectfont
\setlength{\tabcolsep}{0.75pt}
\renewcommand{\arraystretch}{0.88}
\setlength{\aboverulesep}{0.12ex}
\setlength{\belowrulesep}{0.12ex}
\setlength{\cmidrulesep}{0.08ex}

\newcommand{\abscore}[2]{#1{\scriptsize $\pm$#2}}
\newcommand{\abbest}[2]{\textbf{#1}{\scriptsize $\pm$#2}}

\begin{tabular*}{\columnwidth}{
@{\extracolsep{\fill}}
llccccc
@{}
}
\toprule
Policy & Variant & HC & Hop & W2 & Swi & Hum \\
\midrule

\multicolumn{7}{@{}l}{\textit{(a) Target Improvement}}\\[-1.5pt]

\multirow{3}{*}{Gaussian}
& \textbf{Full}
& \abbest{5155}{317}
& \abbest{3289}{263}
& \abbest{4243}{361}
& \abbest{128.7}{8.7}
& \abbest{2364}{423}
\\
& w/o Ref.
& \abscore{1531}{263}
& \abscore{1071}{127}
& \abscore{918}{93}
& \abscore{61.7}{7.8}
& \abscore{887}{157}
\\
& w/o Weight.
& \abscore{3707}{271}
& \abscore{2361}{216}
& \abscore{2238}{319}
& \abscore{90.1}{9.7}
& \abscore{1857}{383}
\\[0.5pt]

\multirow{3}{*}{Diffusion}
& \textbf{Full}
& \abbest{4573}{391}
& \abbest{2632}{337}
& \abbest{3854}{431}
& \abbest{124.9}{9.8}
& \abbest{1462}{263}
\\
& w/o Ref.
& \abscore{1784}{227}
& \abscore{1289}{301}
& \abscore{1930}{349}
& \abscore{52.1}{7.7}
& \abscore{673}{91}
\\
& w/o Weight.
& \abscore{3178}{267}
& \abscore{1945}{359}
& \abscore{2579}{393}
& \abscore{87.4}{8.1}
& \abscore{1086}{169}
\\[0.5pt]

\multirow{3}{*}{Flow}
& \textbf{Full}
& \abbest{4300}{409}
& \abbest{3016}{357}
& \abbest{3650}{451}
& \abbest{124.2}{9.1}
& \abbest{1535}{279}
\\
& w/o Ref.
& \abscore{1947}{331}
& \abscore{1291}{241}
& \abscore{1367}{443}
& \abscore{56.8}{4.2}
& \abscore{669}{241}
\\
& w/o Weight.
& \abscore{3108}{397}
& \abscore{2413}{281}
& \abscore{2597}{503}
& \abscore{86.6}{13.4}
& \abscore{1172}{269}
\\

\midrule

\multicolumn{7}{@{}l}{\textit{(b) Direct Action Fitting}}\\[-1.5pt]

\multirow{3}{*}{Diffusion}
& \textbf{Full}
& \abbest{4573}{391}
& \abbest{2632}{337}
& \abbest{3854}{431}
& \abbest{124.9}{9.8}
& \abbest{1462}{263}
\\
& Native Only
& \abscore{1132}{121}
& \abscore{576}{99}
& \abscore{667}{83}
& \abscore{31.2}{4.9}
& \abscore{467}{71}
\\
& Direct Only
& \abscore{2961}{357}
& \abscore{1938}{319}
& \abscore{2728}{363}
& \abscore{90.2}{8.3}
& \abscore{1082}{211}
\\[0.5pt]

\multirow{3}{*}{Flow}
& \textbf{Full}
& \abbest{4300}{409}
& \abbest{3016}{357}
& \abbest{3650}{451}
& \abbest{124.2}{9.1}
& \abbest{1535}{279}
\\
& Native Only
& \abscore{973}{431}
& \abscore{826}{339}
& \abscore{871}{91}
& \abscore{29.9}{10.8}
& \abscore{376}{219}
\\
& Direct Only
& \abscore{2878}{367}
& \abscore{2044}{292}
& \abscore{2599}{414}
& \abscore{89.8}{7.2}
& \abscore{1108}{231}
\\

\bottomrule
\end{tabular*}
\endgroup
\vspace{-8pt}
\end{table}

We compare Native Only, Direct Only, and Full
(Direct + Native Auxiliary). Native Only fits refined clean
actions through the native diffusion/flow objectives, whereas
Direct Only supervises behavior-anchored action readouts toward
the same targets. As shown in Table~\ref{tab:ablation}, Direct Only achieves
higher mean Final Return than Native Only in both policy families,
while Full obtains the highest mean Final Return among the three
variants on all five tasks. These results support combining direct
action fitting with native auxiliary training.

\paragraph{Target Improvement}
We further ablate the two components of target improvement.
\textit{w/o Refinement} sets $a_i^+=a_i^{\rm exec}$ and retains only
advantage weighting, whereas \textit{w/o Weighting} sets $w_i=1$ and retains
only advantage-guided action refinement.
As shown in Table~\ref{tab:ablation}, removing either component degrades
performance, indicating that advantage weighting and local action refinement
provide complementary contributions to target improvement.

\subsection{Embodied Navigation Post-Training}
\label{sec:navigation}

\begin{table}[!t]
\centering
\caption{Closed-loop wheeled-robot navigation performance on InternScenes.
SR denotes Success Rate and SPL denotes Success weighted by Path Length.}
\label{tab:navigation}
\setlength{\tabcolsep}{4pt}
\begin{tabular}{lcccccc}
\toprule
Method & \multicolumn{2}{c}{Overall} & \multicolumn{2}{c}{Commercial}
& \multicolumn{2}{c}{Home} \\
& SR & SPL & SR & SPL & SR & SPL \\
\midrule
iPlanner & 45.28 & 43.54 & 51.75 & 50.02 & 38.80 & 37.05 \\
ViPlanner & 45.08 & 44.00 & 55.40 & 53.94 & 34.75 & 34.05 \\
NavDP & 70.33 & 66.85 & 72.55 & 69.74 & 68.10 & 63.96 \\
NavOL & 67.00 & 63.00 & 75.20 & 71.10 & 58.80 & 54.90 \\
SIDP & 77.43 & 69.32 & 82.15 & 74.16 & 72.70 & 64.48 \\
NavDP-RL & -- & -- & -- & -- & 68.70 & 62.80 \\
X-NavDP & 88.63 & \textbf{77.67} & 88.55 & \textbf{78.96} & 88.70 & \textbf{76.37} \\
\midrule
\textbf{DR-APO} & \textbf{89.55} & 74.55 & \textbf{89.30} & 76.72 & \textbf{89.80} & 72.38 \\
\bottomrule
\vspace{-15pt}
\end{tabular}
\end{table}

\begin{figure}[!t]
    \centering
    \includegraphics[
        width=\columnwidth
    ]{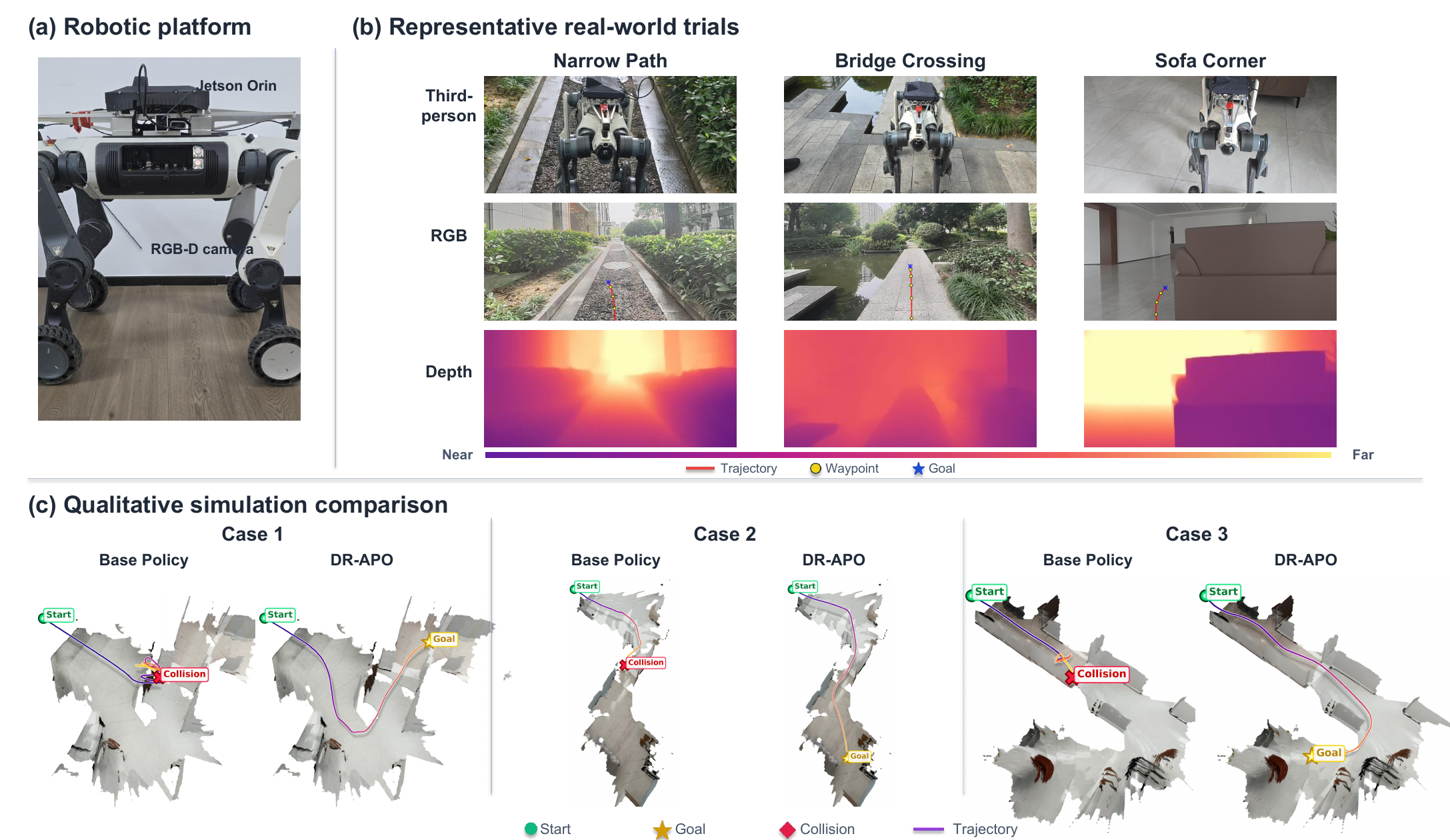}
    \caption{
    Qualitative comparison of pretrained NavDP and DR-APO post-training
in simulated and real-world navigation.
    }
    \label{fig:navigation_qualitative}
    \vspace{-4mm}
\end{figure}

Table~\ref{tab:navigation} compares DR-APO with iPlanner~\cite{iplanner},
ViPlanner~\cite{viplanner}, and diffusion-based navigation baselines.
DR-APO improves pretrained NavDP's overall SR from $70.33\%$ to
$89.55\%$ ($+19.22$ percentage points), exceeding X-NavDP by $0.92$ points.
Its lower SPL ($74.55$ vs.\ $77.67$) highlights a trade-off between
navigation success and path efficiency.
In real-world navigation, DR-APO further improves SR from $25\%$ to $65\%$,
a gain of $40$ percentage points over pretrained NavDP
(Table~\ref{tab:real_navigation}).
Figure~\ref{fig:navigation_qualitative} shows qualitative comparisons in
simulation and the real world.

\begin{table}[t]
\centering
\caption{Real-world navigation success rate (SR) across 20 scenarios.}
\label{tab:real_navigation}
\setlength{\tabcolsep}{8pt}
\begin{tabular}{lcc}
\toprule
Method & NavDP (pretrained) & \textbf{DR-APO} \\
\midrule
SR (\%) & 25 & \textbf{65} \\
\bottomrule
\end{tabular}
\vspace{-4mm}
\end{table}

\section{Conclusion}

DR-APO decouples advantage-guided improvement from action fitting
without likelihood-ratio updates or inference-time critic selection.
On five MuJoCo-v5 tasks, it is competitive with Gaussian SAC and achieves
the best aggregate policy-only results for diffusion and flow policies.
NavDP post-training raises SR from $70.33\%$ to $89.55\%$ in simulation
and from $25\%$ to $65\%$ in real-world navigation.

\bibliographystyle{IEEEtran}
\bibliography{references}

@inproceedings{abdolmaleki2018mpo,
  title     = {Maximum a Posteriori Policy Optimisation},
  author    = {Abdolmaleki, Abbas and Springenberg, Jost Tobias and Tassa, Yuval and Munos, Remi and Heess, Nicolas and Riedmiller, Martin},
  booktitle = {International Conference on Learning Representations},
  year      = {2018},
  url       = {https://arxiv.org/abs/1806.06920}
}

@inproceedings{ajay2023decisiondiffuser,
  title     = {Is Conditional Generative Modeling All You Need for Decision-Making?},
  author    = {Ajay, Anurag and Du, Yilun and Gupta, Abhi and Tenenbaum, Joshua and Jaakkola, Tommi and Agrawal, Pulkit},
  booktitle = {International Conference on Learning Representations},
  year      = {2023},
  url       = {https://arxiv.org/abs/2211.15657}
}

@misc{chen2026pirl,
  title        = {{$\pi_{\mathrm{RL}}$}: Online {RL} Fine-Tuning for Flow-Based Vision-Language-Action Models},
  author       = {Chen, Kang and Liu, Zhihao and Zhang, Tonghe and Guo, Zhen and Xu, Si and Lin, Hao and Zang, Hongzhi and Li, Xiang and Zhang, Quanlu and Yu, Zhaofei and Fan, Guoliang and Huang, Tiejun and Wang, Yu and Yu, Chao},
  year         = {2026},
  note         = {arXiv:2510.25889},
  url          = {https://arxiv.org/abs/2510.25889}
}

@article{chi2024diffusionpolicy,
  title   = {Diffusion Policy: Visuomotor Policy Learning via Action Diffusion},
  author  = {Chi, Cheng and Xu, Zhenjia and Feng, Siyuan and Cousineau, Eric and Du, Yilun and Burchfiel, Benjamin and Tedrake, Russ and Song, Shuran},
  journal = {The International Journal of Robotics Research},
  year    = {2024},
  volume  = {44},
  number  = {10--11},
  pages   = {1684--1704},
  doi     = {10.1177/02783649241273668},
  url     = {https://doi.org/10.1177/02783649241273668}
}

@misc{ding2024qvpo,
  title  = {Diffusion-Based Reinforcement Learning via {Q}-Weighted Variational Policy Optimization},
  author = {Ding, Shutong and Hu, Ke and Zhang, Zhenhao and Ren, Kan and Zhang, Weinan and Yu, Jingyi and Wang, Jingya and Shi, Ye},
  year   = {2024},
  note   = {arXiv:2405.16173},
  url    = {https://arxiv.org/abs/2405.16173}
}

@misc{ding2026genpo,
  title  = {{GenPO}: Generative Diffusion Models Meet On-Policy Reinforcement Learning},
  author = {Ding, Shutong and Hu, Ke and Zhong, Shan and Luo, Haoyang and Zhang, Weinan and Wang, Jingya and Wang, Jun and Shi, Ye},
  year   = {2026},
  note   = {arXiv:2505.18763},
  url    = {https://arxiv.org/abs/2505.18763}
}

@inproceedings{fujimoto2018td3,
  title     = {Addressing Function Approximation Error in Actor-Critic Methods},
  author    = {Fujimoto, Scott and van Hoof, Herke and Meger, David},
  booktitle = {Proceedings of the 35th International Conference on Machine Learning},
  year      = {2018},
  url       = {https://arxiv.org/abs/1802.09477}
}

@inproceedings{haarnoja2018sac,
  title     = {Soft Actor-Critic: Off-Policy Maximum Entropy Deep Reinforcement Learning with a Stochastic Actor},
  author    = {Haarnoja, Tuomas and Zhou, Aurick and Abbeel, Pieter and Levine, Sergey},
  booktitle = {Proceedings of the 35th International Conference on Machine Learning},
  year      = {2018},
  url       = {https://arxiv.org/abs/1801.01290}
}

@inproceedings{ho2020ddpm,
  title     = {Denoising Diffusion Probabilistic Models},
  author    = {Ho, Jonathan and Jain, Ajay and Abbeel, Pieter},
  booktitle = {Advances in Neural Information Processing Systems},
  volume    = {33},
  pages     = {6840--6851},
  year      = {2020}
}

@inproceedings{janner2022diffuser,
  title     = {Planning with Diffusion for Flexible Behavior Synthesis},
  author    = {Janner, Michael and Du, Yilun and Tenenbaum, Joshua B. and Levine, Sergey},
  booktitle = {Proceedings of the 39th International Conference on Machine Learning},
  year      = {2022},
  url       = {https://arxiv.org/abs/2205.09991}
}

@article{internscenes,
  title   = {{InternScenes}: A Large-Scale Simulatable Indoor Scene Dataset with Realistic Layouts},
  author  = {Zhong, Weizhi and Cao, Peng and Jin, Yinqi and others},
  journal = {Advances in Neural Information Processing Systems},
  volume  = {38},
  year    = {2026}
}

@misc{jiang2025adpo,
  title  = {Adaptive Diffusion Policy Optimization for Robotic Manipulation},
  author = {Jiang, Huiyun and Yang, Zhuang},
  year   = {2025},
  note   = {arXiv:2505.08376},
  url    = {https://arxiv.org/abs/2505.08376}
}

@inproceedings{kakade2002approximately,
  title     = {Approximately Optimal Approximate Reinforcement Learning},
  author    = {Kakade, Sham M. and Langford, John},
  booktitle = {International Conference on Machine Learning},
  year      = {2002},
  url       = {https://api.semanticscholar.org/CorpusID:31442909}
}

@inproceedings{lsevpo,
  title     = {Diffusion-Based Reinforcement Learning via {LogSumExp}-Weighted Variational Policy Optimization},
  author    = {Zhang, Rui and An, Jiaming and Wang, Yu and others},
  booktitle = {2025 China Automation Congress (CAC)},
  pages     = {971--976},
  year      = {2025},
  publisher = {IEEE}
}

@inproceedings{lillicrap2016ddpg,
  title     = {Continuous Control with Deep Reinforcement Learning},
  author    = {Lillicrap, Timothy P. and Hunt, Jonathan J. and Pritzel, Alexander and Heess, Nicolas and Erez, Tom and Tassa, Yuval and Silver, David and Wierstra, Daan},
  booktitle = {International Conference on Learning Representations},
  year      = {2016},
  url       = {https://arxiv.org/abs/1509.02971}
}

@inproceedings{lipman2023flowmatching,
  title     = {Flow Matching for Generative Modeling},
  author    = {Lipman, Yaron and Chen, Ricky T. Q. and Ben-Hamu, Heli and Nickel, Maximilian and Le, Matt},
  booktitle = {International Conference on Learning Representations},
  year      = {2023},
  url       = {https://arxiv.org/abs/2210.02747}
}

@misc{ma2025dpmd,
  title  = {Efficient Online Reinforcement Learning for Diffusion Policy},
  author = {Ma, Haitong and Chen, Tianyi and Wang, Kai and Li, Na and Dai, Bo},
  year   = {2025},
  note   = {arXiv:2502.00361},
  url    = {https://arxiv.org/abs/2502.00361}
}

@misc{mark2024parl,
  title  = {Policy Agnostic {RL}: Offline {RL} and Online {RL} Fine-Tuning of Any Class and Backbone},
  author = {Mark, Max Sobol and Gao, Tian and Sampaio, Georgia Gabriela and Srirama, Mohan Kumar and Sharma, Archit and Finn, Chelsea and Kumar, Aviral},
  year   = {2024},
  note   = {arXiv:2412.06685},
  url    = {https://arxiv.org/abs/2412.06685}
}

@misc{mcallister2025fpo,
  title  = {Flow Matching Policy Gradients},
  author = {McAllister, David and Ge, Songwei and Yi, Brent and Kim, Chung Min and Weber, Ethan and Choi, Hongsuk and Feng, Haiwen and Kanazawa, Angjoo},
  year   = {2025},
  note   = {arXiv:2507.21053},
  url    = {https://arxiv.org/abs/2507.21053}
}

@misc{nair2020awac,
  title  = {{AWAC}: Accelerating Online Reinforcement Learning with Offline Datasets},
  author = {Nair, Ashvin and Gupta, Abhishek and Dalal, Murtaza and Levine, Sergey},
  year   = {2020},
  note   = {arXiv:2006.09359},
  url    = {https://arxiv.org/abs/2006.09359}
}

@misc{peng2019awr,
  title  = {Advantage-Weighted Regression: Simple and Scalable Off-Policy Reinforcement Learning},
  author = {Peng, Xue Bin and Kumar, Aviral and Zhang, Grace and Levine, Sergey},
  year   = {2019},
  note   = {arXiv:1910.00177},
  url    = {https://arxiv.org/abs/1910.00177}
}

@inproceedings{peters2007rwr,
  title     = {Reinforcement Learning by Reward-Weighted Regression for Operational Space Control},
  author    = {Peters, Jan and Schaal, Stefan},
  booktitle = {Proceedings of the 24th International Conference on Machine Learning},
  year      = {2007},
  pages     = {745--750},
  doi       = {10.1145/1273496.1273590},
  url       = {https://doi.org/10.1145/1273496.1273590}
}

@misc{psenka2025qsm,
  title  = {Learning a Diffusion Model Policy from Rewards via {Q}-Score Matching},
  author = {Psenka, Michael and Escontrela, Alejandro and Abbeel, Pieter and Ma, Yi},
  year   = {2025},
  note   = {arXiv:2312.11752},
  url    = {https://arxiv.org/abs/2312.11752}
}

@misc{ren2024dppo,
  title  = {Diffusion Policy Policy Optimization},
  author = {Ren, Allen Z. and Lidard, Justin and Ankile, Lars L. and Simeonov, Anthony and Agrawal, Pulkit and Majumdar, Anirudha and Burchfiel, Benjamin and Dai, Hongkai and Simchowitz, Max},
  year   = {2024},
  note   = {arXiv:2409.00588},
  url    = {https://arxiv.org/abs/2409.00588}
}

@inproceedings{schulman2015trpo,
  title     = {Trust Region Policy Optimization},
  author    = {Schulman, John and Levine, Sergey and Moritz, Philipp and Jordan, Michael I. and Abbeel, Pieter},
  booktitle = {Proceedings of the 32nd International Conference on Machine Learning},
  year      = {2015},
  url       = {https://proceedings.mlr.press/v37/schulman15.html}
}

@misc{schulman2017ppo,
  title  = {Proximal Policy Optimization Algorithms},
  author = {Schulman, John and Wolski, Filip and Dhariwal, Prafulla and Radford, Alec and Klimov, Oleg},
  year   = {2017},
  note   = {arXiv:1707.06347},
  url    = {https://arxiv.org/abs/1707.06347}
}

@misc{shi2026flowdpg,
  title  = {{FlowDPG}: Deterministic Policy Gradient on Flow Matching Policies for Real-World Manipulation},
  author = {Shi, Kexin and Shi, Junyao and Hebbar, Poorvi and Zhao, Zhuolun and Amarnath, Tarun and Su, Yifan and Bahl, Shikhar and Pathak, Deepak},
  year   = {2026},
  note   = {arXiv:2606.22303},
  url    = {https://arxiv.org/abs/2606.22303}
}

@inproceedings{todorov2012mujoco,
  title     = {{MuJoCo}: A Physics Engine for Model-Based Control},
  author    = {Todorov, Emanuel and Erez, Tom and Tassa, Yuval},
  booktitle = {2012 IEEE/RSJ International Conference on Intelligent Robots and Systems},
  year      = {2012},
  pages     = {5026--5033},
  publisher = {IEEE},
  doi       = {10.1109/IROS.2012.6386109},
  url       = {https://doi.org/10.1109/IROS.2012.6386109}
}

@inproceedings{wang2020crr,
  title     = {Critic Regularized Regression},
  author    = {Wang, Ziyu and Novikov, Alexander and Zolna, Konrad and Springenberg, Jost Tobias and Reed, Scott and Shahriari, Bobak and Siegel, Noah and Merel, Josh and Gulcehre, Caglar and Heess, Nicolas and de Freitas, Nando},
  booktitle = {Advances in Neural Information Processing Systems},
  year      = {2020},
  url       = {https://arxiv.org/abs/2006.15134}
}

@inproceedings{wang2023diffusionql,
  title     = {Diffusion Policies as an Expressive Policy Class for Offline Reinforcement Learning},
  author    = {Wang, Zhendong and Hunt, Jonathan J. and Zhou, Mingyuan},
  booktitle = {International Conference on Learning Representations},
  year      = {2023},
  url       = {https://arxiv.org/abs/2208.06193}
}

@misc{wang2024dacer,
  title  = {Diffusion Actor-Critic with Entropy Regulator},
  author = {Wang, Yinuo and Wang, Likun and Jiang, Yuxuan and Zou, Wenjun and Liu, Tong and Song, Xujie and Wang, Wenxuan and Xiao, Liming and Wu, Jiang and Duan, Jingliang and Li, Shengbo Eben},
  year   = {2024},
  note   = {arXiv:2405.15177},
  url    = {https://arxiv.org/abs/2405.15177}
}

@misc{rfm,
  title  = {Reverse Flow Matching: A Unified Framework for Online Reinforcement Learning with Diffusion and Flow Policies},
  author = {Li, Zichao and Tang, Shisong and Azizan, Navid},
  year   = {2026},
  note   = {arXiv:2601.08136},
  url    = {https://arxiv.org/abs/2601.08136}
}

@misc{yang2023dipo,
  title  = {Policy Representation via Diffusion Probability Model for Reinforcement Learning},
  author = {Yang, Long and Huang, Zhixiong and Lei, Fenghao and Zhong, Yucun and Yang, Yiming and Fang, Cong and Wen, Shiting and Zhou, Binbin and Lin, Zhouchen},
  year   = {2023},
  note   = {arXiv:2305.13122},
  url    = {https://arxiv.org/abs/2305.13122}
}

@misc{iplanner,
  title  = {{iPlanner}: Imperative Path Planning},
  author = {Yang, Fan and Wang, Chuan and Cadena, Cesar and others},
  year   = {2023},
  note   = {arXiv:2302.11434},
  url    = {https://arxiv.org/abs/2302.11434}
}

@misc{navdp,
  title  = {{NavDP}: Learning Sim-to-Real Navigation Diffusion Policy with Privileged Information Guidance},
  author = {Cai, Wenzhe and Peng, Jiaqi and Yang, Yuqiang and others},
  year   = {2025},
  note   = {arXiv:2505.08712},
  url    = {https://arxiv.org/abs/2505.08712}
}

@misc{navdprl,
  title  = {Beyond Imitation: Reinforcement Learning Fine-Tuning for Adaptive Diffusion Navigation Policies},
  author = {Sheng, Jiawei and Bai, Rui and Xu, Kai and others},
  year   = {2026},
  note   = {arXiv:2603.12868},
  url    = {https://arxiv.org/abs/2603.12868}
}

@misc{navol,
  title  = {{NavOL}: Navigation Policy with Online Imitation Learning},
  author = {Wei, Xinyu and Gu, Chenghao and Zhang, Lei},
  year   = {2026},
  note   = {arXiv:2605.11762},
  url    = {https://arxiv.org/abs/2605.11762}
}

@inproceedings{viplanner,
  title     = {{ViPlanner}: Visual Semantic Imperative Learning for Local Navigation},
  author    = {Roth, Pascal and Nubert, Julian and Yang, Fan and others},
  booktitle = {2024 IEEE International Conference on Robotics and Automation (ICRA)},
  pages     = {5243--5249},
  year      = {2024},
  publisher = {IEEE}
}

@misc{yang2026xnavdp,
  title  = {{X-NavDP}: Generalizing Navigation Diffusion Policy to Novel Behavior and Embodiments with Group {Q}-Score Reweighted Matching},
  author = {Yang, Tianyu and Zeng, Yiming and Cai, Wenzhe and Yang, Yuqiang and Peng, Jiaqi and Cheng, Hui and Pang, Jiangmiao and Wang, Tai},
  year   = {2026},
  note   = {arXiv:2607.28560},
  url    = {https://arxiv.org/abs/2607.28560}
}

@misc{zhang2026reinflow,
  title  = {{ReinFlow}: Fine-Tuning Flow Matching Policy with Online Reinforcement Learning},
  author = {Zhang, Tonghe and Yu, Chao and Su, Sichang and Wang, Yu},
  year   = {2026},
  note   = {arXiv:2505.22094},
  url    = {https://arxiv.org/abs/2505.22094}
}

@misc{zhang2026sidp,
  title  = {Self-Imitated Diffusion Policy for Efficient and Robust Visual Navigation},
  author = {Zhang, Runhua and Hou, Junyi and Cheng, Changxu and Chen, Qiyi and Wang, Tao and Zhao, Wuyue},
  year   = {2026},
  note   = {arXiv:2601.22965},
  url    = {https://arxiv.org/abs/2601.22965}
}

\end{document}